\documentclass[runningheads]{llncs}
\usepackage{graphicx}

\usepackage{amsmath}
\DeclareMathOperator*{\argmax}{arg\,max}

\usepackage{amsfonts}

\usepackage{booktabs}
\usepackage{multirow}
\usepackage{colortbl}
\usepackage{xcolor}

\usepackage{marvosym}

\usepackage{hyperref}
\hypersetup{
    linkbordercolor=white,  
    urlbordercolor=white
}

\definecolor{f}{rgb}{0.9, 0.9, 0.0}
\definecolor{s}{rgb}{0.8, 0.8, 0.8}

\usepackage{cleveref}     

\begin{document}
\title{LabelGS: Label-Aware 3D Gaussian Splatting for 3D Scene Segmentation}

%
%
\author{Yupeng Zhang\inst{1} \and
Dezhi Zheng\inst{1} \and
Ping Lu\inst{2} \and
Han Zhang\inst{2} \and
Lei Wang\inst{2} \and
Liping Xiang \inst{2} \and
Cheng Luo \inst{3} \and
Kaijun Deng \inst{1} \and
Xiaowen Fu \inst{1} \and
Linlin Shen \inst{1,4,5}\textsuperscript{(\Letter)} \and
Jinbao Wang \inst{4,5}
}

%
%
 \institute{
 College of Computer Science and Software Engineering, Shenzhen University, China\\
 \and
 ZTE Co., Ltd, China\and
 King Abdullah University of Science and Technology, Saudi Arabia \and
 Computer Vision Institute, School of Artifcial Intelligence, Shenzhen University, Shenzhen, China\and
 Guangdong Provincial Key Laboratory of Intelligent Information Processing 
 \email{zhangyupeng2022@email.szu.edu.cn}
 \email{llshen@szu.edu.cn}
 }
\maketitle              
\begin{abstract}
3D Gaussian Splatting (3DGS) has emerged as a novel explicit representation for 3D scenes, offering both high-fidelity reconstruction and efficient rendering. However, 3DGS lacks 3D segmentation ability, which limits its applicability in tasks that require scene understanding. The identification and isolating of specific object components is crucial. To address this limitation 
, we propose Label-aware 3D Gaussian Splatting (LabelGS), a method that augments the Gaussian representation with object label.
LabelGS introduces cross-view consistent semantic masks for 3D Gaussians and employs a novel Occlusion Analysis Model to avoid overfitting occlusion during optimization, Main Gaussian Labeling model to lift 2D semantic prior to 3D Gaussian and Gaussian Projection Filter to avoid Gaussian label conflict. 
Our approach achieves effective decoupling of Gaussian representations and refines the 3DGS optimization process through a random region sampling strategy, significantly improving efficiency. Extensive experiments demonstrate that LabelGS outperforms previous state-of-the-art methods, including Feature-3DGS, in the 3D scene segmentation task. Notably, LabelGS achieves a remarkable \textbf{22}$\times$ speedup in training compared to Feature-3DGS, at a resolution of $1440\times1080$. Our code will be at \url{https://github.com/garrisonz/LabelGS}.

\end{abstract}
\section{Introduction}

3D scene segmentation is a fundamental task for understanding and interpreting complex 3D environments. This capability is crucial for a wide range of applications, including 3D medical data analysis~\cite{taha2015metrics}, 3D semantic understanding and manipulation~\cite{tchapmi2017segcloud}, robotic navigation~\cite{guan2022ga}, autonomous driving~\cite{zhou2020joint}, and beyond.
The challenge of 3D scene segmentation has long been hampered by the scarcity of semantically annotated 3D data. 
To avoid this limitation, researchers~\cite{qin2023langsplat,zhou2024feature,gaussian_grouping} have used 2D image prior as supervise, and segment 2D feature map after projection of Gaussians.
Although innovative, this 2D-to-3D transfer approach faces two critical challenges. i) First, it struggles to capture the 3D spatial nature of scenes after the projection process. 
ii) Second, the computational cost of learning high-dimension semantic feature of 3D Gaussian is expensive. 
These issues highlight the need for a computationally efficient method that segment directly 3D Gaussians.

\begin{figure}[t]
  \centering
  \includegraphics[width=0.85\linewidth]{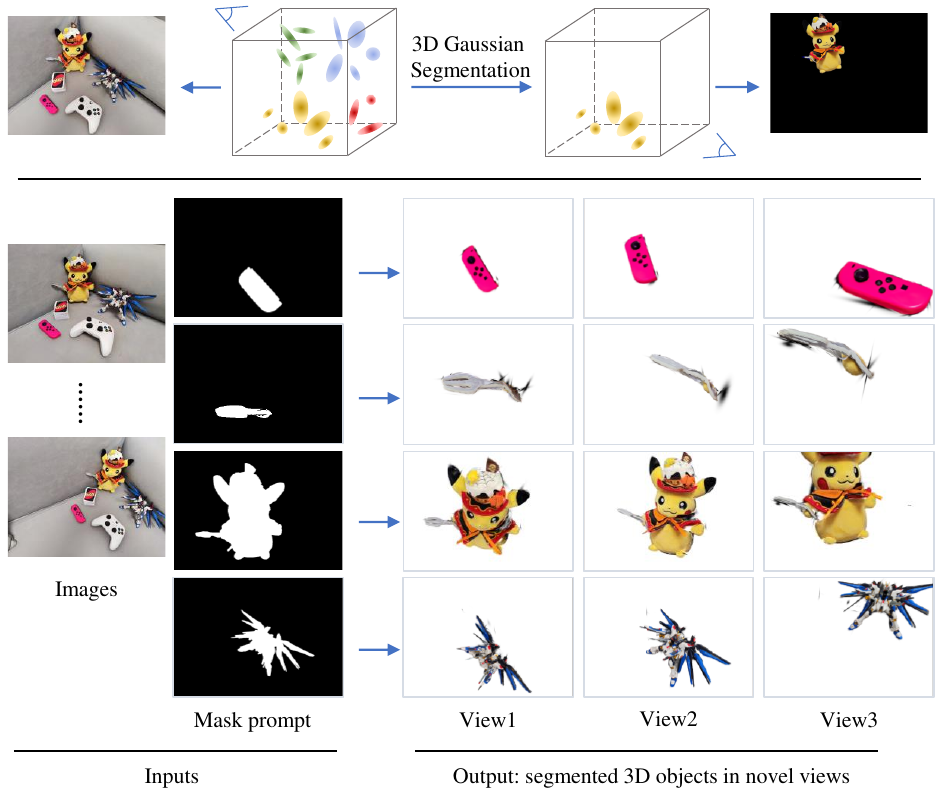}
  \caption{
  Illustration of segmented 3D objects obtained by our method (LabelGS).
  }
  \label{fig:task}
\end{figure}



We propose Label-aware 3D Gaussian Splatting~\cite{kerbl3Dgaussians}~(LabelGS) for 3D scene segmentation, a straightforward yet effective method for assigning object labels to individual Gaussians. 
Our approach consists of several key aspects. 
i) We obtain cross-view masks by DEVA~\cite{cheng2023deva} model, and generate occlusion relations between masks by Occlusion Analysis Model.
ii) We adopt Main Gaussian Labeling strategy to lift 2D label to 3D Gaussian.
iii) We introduce a Gaussian Projection Filter~(GPF) to resolve conflicts of multiple labels assigned to a same 3D Gaussian.  

Overall, the main contributions of this paper are summarized as follows:
\begin{itemize}
\item 
We propose 
a pipeline of 3D segmentation labeling in a 3DGS scene.

\item 
We 
introduce Occlusion Analysis Model to improve 3D object representation, Main Gaussian Labeling to support 3D segmentation.

\item
A new benchmark for 3D scene segmentation is proposed, and experimental results show that our method outperforms the state-of-the-art methods on the 3D segmentation task, i.e. \textbf{22}$\times$ faster than Feature-3DGS~\cite{zhou2024feature} at 1440 $\times$ 1080 resolution.
\end{itemize}

\section{Related Work}
Inspired by the success of Segment Anything Model (SAM)~\cite{kirillov2023segany}, numerous works have extended SAM's 2D segmentation capabilities to NeRF~\cite{lerf2023,mueller2022instant,barron2021mip,tancik2022block} for 3D segmentation~\cite{lerf2023,ying2023omniseg3d,cen2023sa3d}.
For instance, SA3D~\cite{cen2023sa3d} uses SAM for mask inverse rendering and cross-view self-prompting to predict the 3D mask of the target object.
OmniSeg3D~\cite{ying2023omniseg3d} lifts multi-view inconsistent 2D segmentations from SAM into a consistent 3D feature field to achieve 3D segmentation.
Although these works can achieve segmentation results in rendering, their 3D representations are based on MLPs, which are 3D graphic primitives~\cite{mueller2022instant} that cannot be segmented.

3DGS uses a set of 3D Gaussians to represent a 3D scene, with each 3D Gaussian serving as a graphical primitive, facilitating the segmentation of 3D primitives.
Feature-3DGS lifts Lseg~\cite{li2022languagedriven} 2D feature semantic information to 3D Gaussians, for 3D Gaussian segmentation.
LangSplat~\cite{qin2023langsplat} introduces an auto-encoder\cite{kingma2013auto,hinton1993autoencoders} to compress semantic information, improving modeling speed.
Gaussian Grouping~\cite{gaussian_grouping} uses a video tracking model~\cite{cheng2023deva} to obtain consistent 2D segmentation, guiding 3D Gaussians to cluster or disperse.
Unlike these methods of learning the segmentation information, our approach directly lifts the consistent 2D segmentation results to 3D Gaussians, achieving 3D Gaussian segmentation with precise boundaries.

\section{Method}

\begin{figure*}[tb]
  \centering
  \includegraphics[width=1\textwidth]{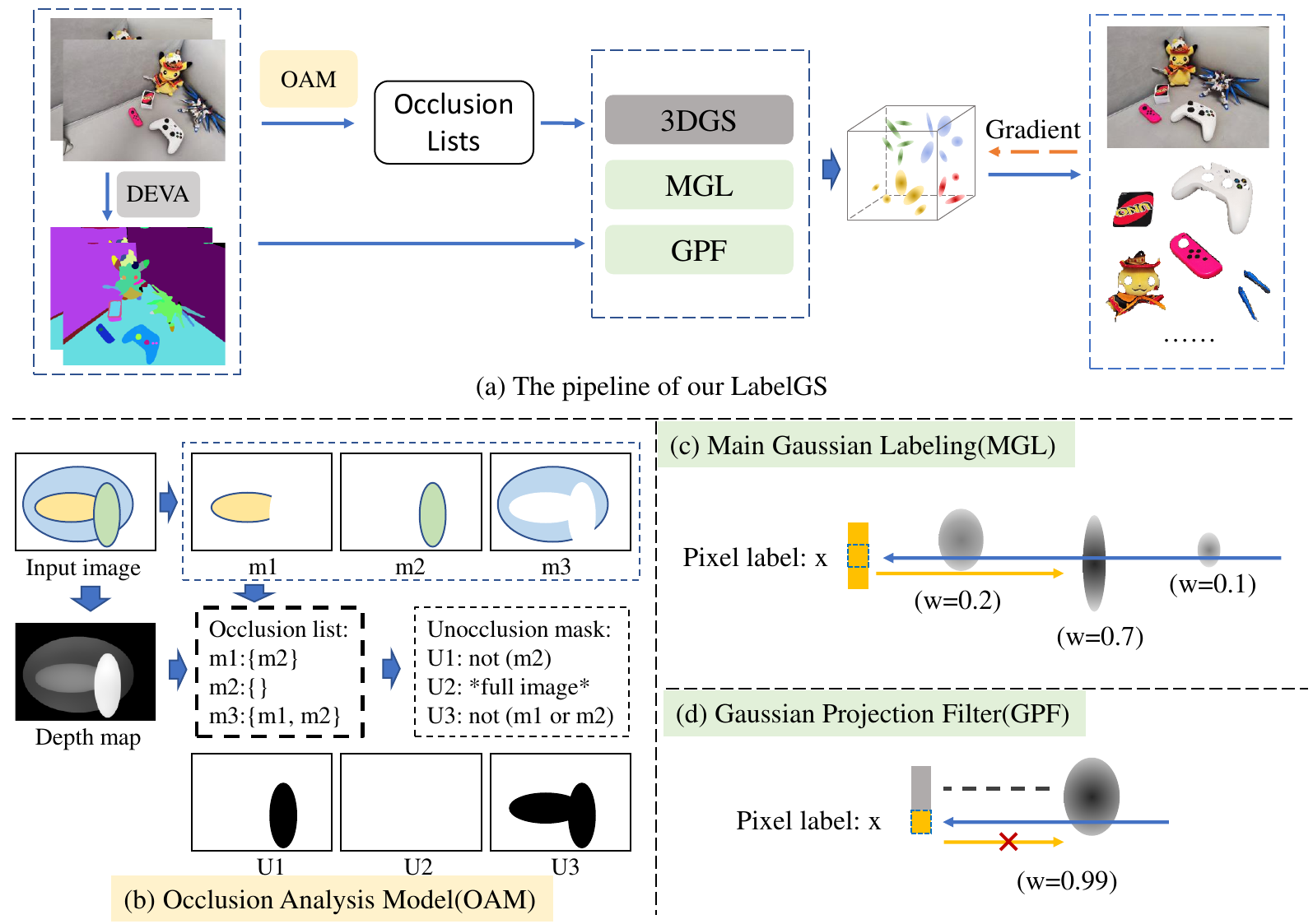}
  \caption{The framework of LabelGS.
  We obtain cross-view masks from \emph{DEVA}, occlusion relationship between masks by Occlusion Analysis Model, and lift these 2D pixel labels to 3D Gaussians by Main Gaussian Labeling and Gaussian Projection Filter. 
  }
  \label{fig:pipeline1}
\end{figure*}

In this section, we elaborate on how our LabelGS assigns a label to the 3D Gaussian so that optimizing 3D scene at object levels and achieving 3D scene segmentation. Fig.~\ref{fig:pipeline1} depicts the framework of our proposed LabelGS.

\subsection{Cross-view mask}

\begin{figure*}[t]
  \centering
  \includegraphics[width=0.75\textwidth]{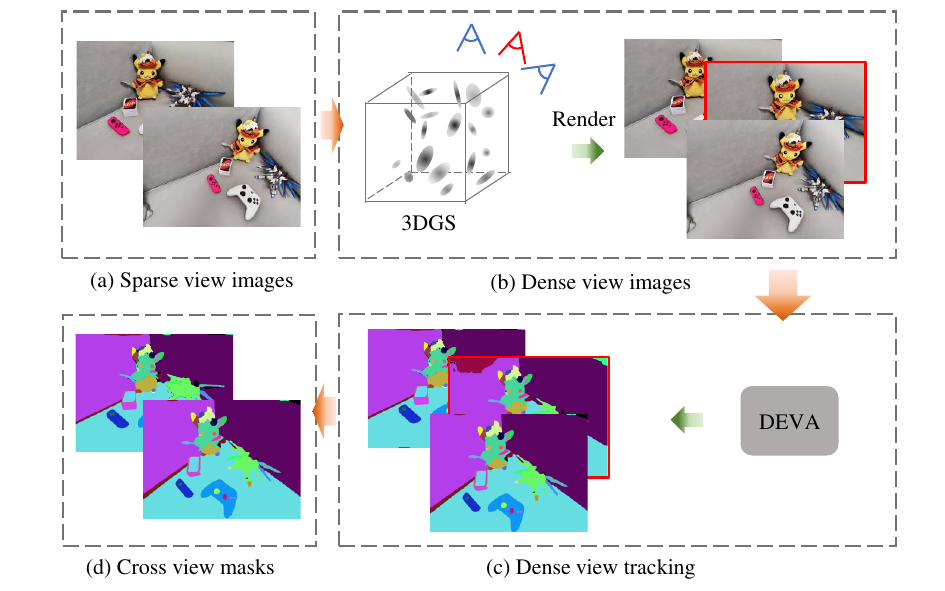} 
  \caption{Densification of view(DOV) for sparse view input images.
  For the input images (a), we model the native 3DGS to obtain dense view images, as video frames (b).
  Next, 2D cross-view masks (c) are generated by DEVA~\cite{cheng2023deva}.
  (d) We retain the tracking results corresponding to the input image. }
  \label{fig:preprocess}
\end{figure*}

Given images $I = \{ I_1, I_2, ..., I_t \}$, DEVA outputs the cross-view consistent masks of each image: $M^i = \{ M_1^i, M_2^i, ..., M_N^i \}, i\in [1, t]$, where $N$ is the number of overall labels, and $M^i_j$ represents the mask of the $j$-th label in the $i$-th image.
To obtain more dense view images, we first train a vanilla 3DGS and then render from pseudo-camera views. The pseudo-camera views are the insertion views between each two adjacent views in the training images. This view-densifying operation is denoted as $D(\cdot)$.
Then, we only retain the masks of the frames that appear in the training set, denoted as $E(\cdot)$. 
The whole preprocess is formulated as follows:

\begin{equation}
M = E(S(D(I))).
\end{equation}
where $M=\{M^1, M^2, ..., M^t\}$ is the cross-view consistent masks corresponding to input images, referring to Fig.~\ref{fig:preprocess}.

With these cross-view consistent masks, we optimize 3D Gaussians for each instance and the whole scene simultaneously.
However, for an object occluded by others, its masks may mislead the 3D Gaussian model into overfitting the occlusion, resulting in an incomplete 3D object.
To address this issue, we introduce an Occlusion Analysis Model to skip occlusion affect.

\subsection{Occlusion Analysis Model} 

The unocclusion mask is used to exclude the occluded areas from loss computation, so that 3D information on these areas learns from other visible views, instead of overfitting the empty space of occlusion. The illustration of the unocclusion mask is shown in Fig.~\ref{fig:pipeline1}.
Given an input image and its cross-view masks, the monocular depth estimation model, i.e. DepthAnythingV2~\cite{yang2024depth}, predicts depth estimation for each pixel. For a segmentation mask $M_k$, we identify an adjacent mask $M_{k'}$ and compute the average depth of two adjacent boundary segments, denoted as $D_{k,k'}(k)$ for $M_k$ and $D_{k,k'}(k')$ for $M_{k'}$. If $D_{k, k'}(k')$ is smaller than $D_{k, k'}(k)$, $M_{k'}$ is added to the occlusion list of $M_k$, denoted as $O_k$. The occlusion list $O_k$ contains all masks that occlude $M_k$ and is defined as:
\begin{equation}
O_k = \{M_j | D_{k,j}(j) < D_{k,j}(k)\}.
\end{equation}
The unoccluded mask of $M_k$ is the remaining area after removing the occluded mask list $O_k$, expressed as:
\begin{equation}
U_k = \neg(M_{j1} \lor M_{j2} \lor ... \lor M_{j|O_k|}).
\end{equation}
Similarly, the unoccluded mask for the $k$-th segmentation mask from the view $i$ is denoted as $U^i_k$.

\subsection{Label-Aware 3D Gaussian Splatting}

Having obtained the cross-view masks and the mask occlusion lists of the 2D input image $I$, we can learn a label-aware 3D scene by modeling the relations between 3D Gaussain and 2D pixels.
We represent the label-aware 3D scene based on 3D Gaussian Splatting due to its efficient rendering.

\subsubsection{3D Gaussian Splatting}
A 3D scene can be represented as a set of 3D Gaussian primitives $\{G_j | j = 1, ..., P\}$, with each 3D Gaussian $G_j$ defined by a mean vector $\mu_j \in \mathbb{R}^3$ and 
a covariance matrix $\Sigma_j$.
The radiance in 3D point $x$ is defined as:
\begin{equation}
G_j(x) = exp(-\frac{1}{2}(x-\mu_j)^T \Sigma_j^{-1}(x-\mu_j)).
\end{equation}
Each pixel is obtained by alpha-blending technology after linearly approximating the 3D Gaussians to 2D Gaussians. Mathematically, the rendered color at pixel $v$ is:
\begin{equation}
C(v) = \sum_{j\in N_{tile}}c_j\alpha_j T_j,
\quad
T_j = \prod_{i=1}^{j-1}(1-\alpha_i).
\end{equation}
Here, $N_{tile}$ denotes the Gaussians in the tile, $c_j$ is the color of the $j$-th Gaussian, and $\alpha_i = o_jG_j^{2D}(v)$.
$o_j$ denotes the opacity of the $j$-th Gaussian and $G_j^{2D}(v)$ is the projection function of the j-th Gaussian onto the pixel $v$ in the 2D image.

\subsubsection{Main Gaussain Labeling}
We propose label-aware Gaussian Splatting, where a label attribute is added to each 3D Gaussian.
The label attribute of the $j$-th Gaussian is denoted as $B_j$, and the entire Gaussian model is represented as the set $G= \{ (G_j, B_j) | j = 1, ..., P \}$.
These labels are obtained by tracking the training images using DEVA.
The augmented Gaussians are named as 3D label Gaussians.
During training, for the $i$-th viewpoint, the label of pixel $v$ is assigned to the Gaussian that contributes the most to the rendering, as shown in Fig.~\ref{fig:pipeline1}(b):  
\begin{equation}
    \begin{aligned}
        B_j = L^i(v), \ 
        \text{s.t.} \ j = \argmax_k (\alpha_k T_k),
    \end{aligned}
\end{equation}
where $L^i(v)$ denotes the label at position $v$ in the image from view $i$, $j$ represents the Gaussian index that contributes the most to the color of the pixel $v$ during rendering.
By embedding the label directly into the Gaussian, we obtain labeled Gaussians.

\subsubsection{Gaussian Projection Filter}
When one 3D Gaussian is splatted onto a 2D plane, the pixels within the Gaussian splatting projection can belong to different label areas, meaning that the same Gaussian may become the primary contributor to multiple pixels with different labels, leading to Gaussian label conflicts.
To address these conflicts, we introduce the Gaussian projection filter (GPF), as shown in Fig.~\ref{fig:pipeline1}(c).
This filter ensures that a label is lifted to a 3D Gaussian only if the central projection of the Gaussian and the current pixel position belong to the same label region.
Mathematically, the lifting operation is as follows:

\begin{equation}
\begin{aligned}
B_j = L^i(v), \ 
\text{s.t.}\ j = {\argmax_i (\alpha_iT_i)}, \  
L^i(v) = L^i(Proj_i(\mu_j)),
\end{aligned}
\end{equation}
where $Proj_i(.)$ denotes the projection function from 3D to 2D in view $i$.


%
\subsubsection{Optimization.}
We optimize the Gaussian model using consistent masks. The loss function is defined as:
\begin{equation}
L_{label} = \sum_{i=1}^{t} \sum_{k=1}^{N} L_1(Proj_i(H(G,k)) U^i_k, I^i M^i_k U^i_k)
\end{equation}
where 
\( L_1 \) is the \( L_1 \) loss, 
\( H(G,k) \) denotes all Gaussians with the label \( k \) in the Gaussian model, and \( M^i_k \) and \( U^i_k \) represent the \( k \)-th mask and the corresponding unoccluded mask from the \( i \)-th viewpoint, respectively.
The set \( H(G,k) \) is given by:
\begin{equation}
H(G,k) = \{ G_j \mid B_j = k, j \in [1, P] \},
\end{equation}
where \( B_j \) is the label of the \( j \)-th Gaussian, and \( P \) is the total number of Gaussians in the model.
We also employ the full image loss, which includes both the \( L_1 \) loss and the \( L_{ssim} \) loss, as defined in 3DGS.
Thus, the overall loss function is:

\begin{equation}
Loss = (1-\lambda_1) L_1 + \lambda_1 L_{ssim} + \lambda_2 L_{label}.
\end{equation}

\begin{figure}[t]
  \centering
  \includegraphics[width=0.6\textwidth]{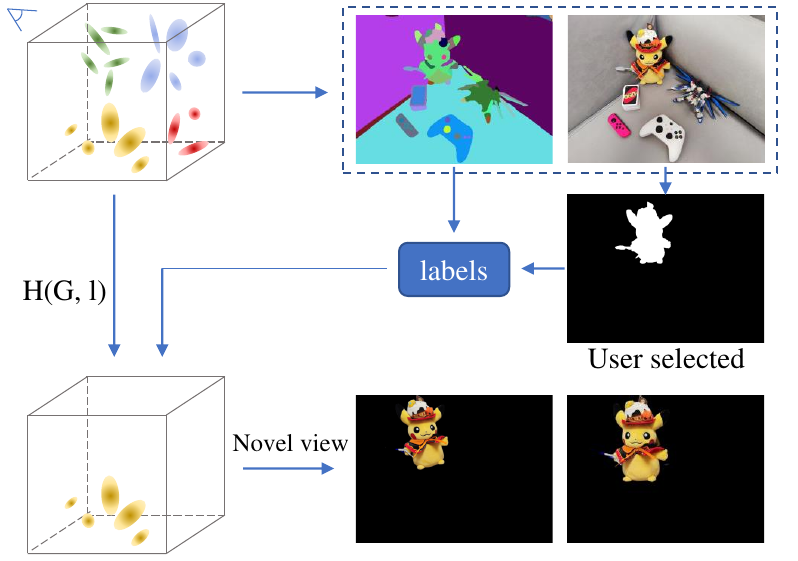}
  \caption{The inference process of our LabelGS.
  The optimized 3DGS renders the image and the label map from a given viewpoint.
  Users select the target mask to obtain the labels, which is used to extract 3D Gaussians.}
  \label{fig:inference}
\end{figure}

\subsubsection{Inference}
The Gaussian model renders both an image and a 2D label map $L^i$ from a view $i$.
The user selects a mask $M_p$ based on the rendered image.
After applying the mask on the label map, we obtain a set of labels $l = \Phi(L^iM_p)$.
These labels are used to extract Gaussians with the same label from the Gaussian model, denoted as $H(G, l)$.
These extracted Gaussians render a novel view image of the target object, shown as Fig.\ref{fig:inference}.

\section{Experiments}

\subsection{Experimental Setup}

\begin{table}[tb]
  \caption{Number of training and test images for each dataset.}
  \label{tab:dataset}
  \centering
 \begin{tabular}{l*{10}{c}}
    \toprule
    Item & 3D-OVS &  LLFF & LERF &  Mip-NeRF360 \\
    \midrule
    Training images & 289 & 278 & 612 & 1285 \\
    Test images & 40 & 27 & 6 & 22 \\
    Test objects & 160 & 47 & 35 & 60\\
    \bottomrule
  \end{tabular}
\end{table}

\begin{table*}[tb]
\centering
\caption{3D segmentation quality. The mIoU and PSNR of rendered images from extracted Gaussians in novel views are reported.}
\label{tab:comparison}
\resizebox{1\textwidth}{!}
{
\begin{tabular}{lcccccccc}
\toprule
\textbf{dataset} & \multicolumn{2}{c}{\textbf{3D-OVS}} & \multicolumn{2}{c}{\textbf{LLFF}} & \multicolumn{2}{c}{\textbf{LERF}} & \multicolumn{2}{c}{\textbf{Mip-NeRF360}} \\ 
\cmidrule(lr){2-3} \cmidrule(lr){4-5} \cmidrule(lr){6-7} \cmidrule(lr){8-9}
\textbf{Method} & mIoU~$(\uparrow)$ & PSNR~$(\uparrow)$ & mIoU~$(\uparrow)$ & PSNR~$(\uparrow)$ & mIoU~$(\uparrow)$ & PSNR~$(\uparrow)$ & mIoU~$(\uparrow)$ & PSNR~$(\uparrow)$ \\ 
\midrule
Feature-3DGS\cite{zhou2024feature}    & 0.22 & 19.7 & - & - & - & - & - & - \\
LangSplat\cite{qin2023langsplat}       & 0.11 & 19.3 & 0.36 & 17.3 & 0.13 & 19.51 & 0.12 & 20.8  \\
Gaussian Grouping\cite{gaussian_grouping}  & 0.57 & 23.0 & 0.51  & 18.7  & 0.33 & 16.9 & 0.52 &  21.8 \\
SAGS\cite{hu2024semantic} & 0.76 & 30.4 & 0.82 & 27.1 & 0.57 & 27.6& 0.57 & 25.6\\

\midrule
LabelGS w/o DOV  & 0.922       & 33.99 & - & - & - & - & - & - \\
LabelGS w/o GPF  & \cellcolor{s}0.926 & 34.09              & \cellcolor{s}0.848 & 27.97              & \cellcolor{f}0.82 & \cellcolor{f}33.30 & \cellcolor{s}0.76 & \cellcolor{f}32.73\\
LabelGS w/o OAM  & \cellcolor{f}0.928 & \cellcolor{s}34.19 & 0.844              & \cellcolor{s}28.01 & \cellcolor{s}0.80 & \cellcolor{s}32.92 & \cellcolor{f}0.79 & \cellcolor{s}32.47\\
LabelGS          & 0.925              & \cellcolor{f}34.26 & \cellcolor{f}0.851 & \cellcolor{f}28.02 & 0.75              & 32.47              & 0.75              & 32.42\\
\bottomrule
\end{tabular}}
\end{table*}

\begin{table*}[tb]
\centering
\caption{3D segmentation quality. The SSIM and LPIPS of rendered images from extracted Gaussians in novel views are reported.}
\label{tab:comparison_ssim_lpips}
\resizebox{1\textwidth}{!}
{
\begin{tabular}{lcccccccc}
\toprule
\textbf{dataset} & \multicolumn{2}{c}{\textbf{3D-OVS}} & \multicolumn{2}{c}{\textbf{LLFF}} & \multicolumn{2}{c}{\textbf{LERF}} & \multicolumn{2}{c}{\textbf{Mip-NeRF360}} \\ 
\cmidrule(lr){2-3} \cmidrule(lr){4-5} \cmidrule(lr){6-7} \cmidrule(lr){8-9}
\textbf{Method} & SSIM~$(\uparrow)$ & LPIPS~$(\downarrow)$ & SSIM~$(\uparrow)$ & LPIPS~$(\downarrow)$ & SSIM~$(\uparrow)$ & LPIPS~$(\downarrow)$ & SSIM~$(\uparrow)$ & LPIPS~$(\downarrow)$ \\ 
\midrule
Feature-3DGS\cite{zhou2024feature} & 0.77 & 0.22 & - & - & - & - & - & -  \\
LangSplat\cite{qin2023langsplat}   & 0.8  & 0.17 & 0.75 & 0.26 & 0.72 & 0.24 & 0.78 & 0.19 \\
Gaussian Grouping\cite{gaussian_grouping} & 0.91 & 0.07 & 0.81 & 0.18 & 0.72 & 0.23 & 0.86 & 0.14 \\
SAGS\cite{hu2024semantic} & \textbf{0.98} & \textbf{0.02} & 0.95 & \textbf{0.06}& 0.93 & 0.07 & 0.95 & \textbf{0.04} \\
\textbf{LabelGS(ours)} & \textbf{0.98} & \textbf{0.02} & \textbf{0.93}& 0.07 & \textbf{0.97} & \textbf{0.04}& \textbf{0.96} & \textbf{0.04}\\
\bottomrule
\end{tabular}}
\end{table*}

\begin{table}[tb]
  \caption{Comparisons of training time for 15,000 iterations on the \emph{sofa} scene from the 3D-OVS dataset.}
  \centering
  \begin{tabular}{@{}lc@{}}
  \toprule
    \textbf{Method} & \textbf{minutes~($\downarrow$)}\\
  \midrule
    Feature-3DGS\cite{zhou2024feature} & 443 \\
    Gaussian Grouping\cite{gaussian_grouping} & 28 \\
    LangSplat\cite{qin2023langsplat} & 22 \\
    \textbf{LabelGS (Ours)} &\textbf{20} \\
  \bottomrule
  \end{tabular}
  \label{tab:speed}
\end{table}

\begin{figure*}[t]
  \centering
  \includegraphics[width=1\textwidth]{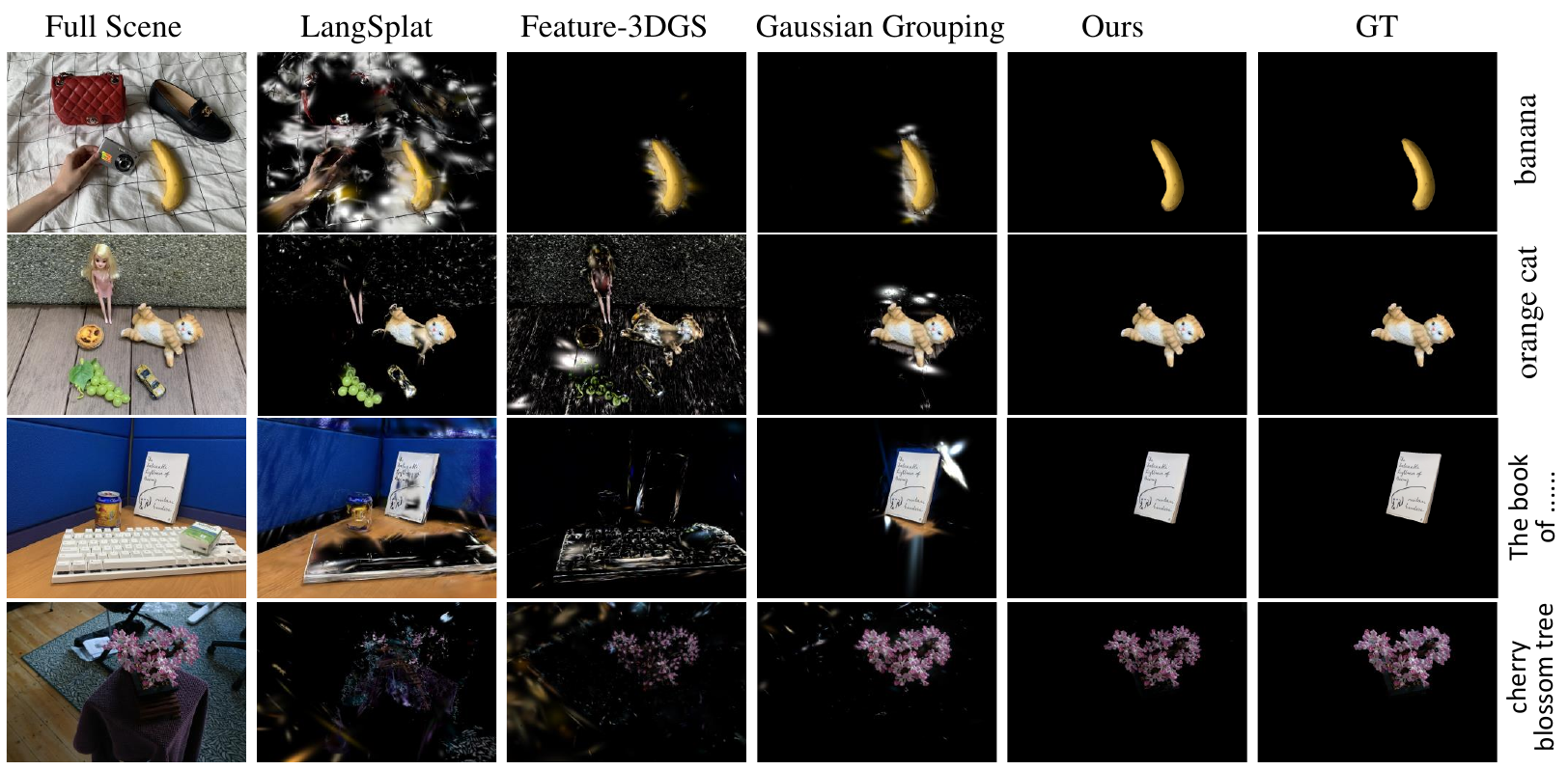}
  \caption{Qualitative comparisons of 3D Gaussian segmentation in test views. We query the text and its mask in the training view.}
  \label{fig:compare}
\end{figure*}

\begin{figure*}[t]
  \centering
  \includegraphics[width=1\textwidth]{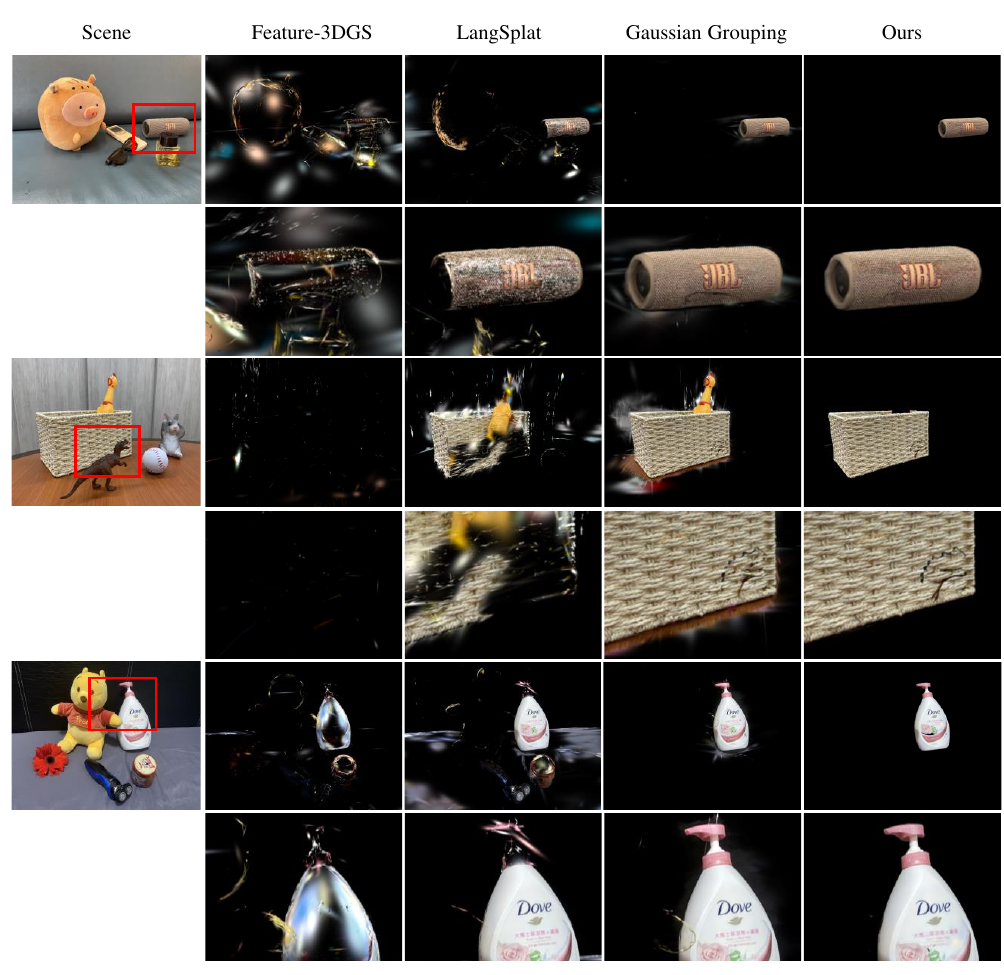}
  \caption{The comparisons of occlusion area reconstruction after applying gaussian segmentation.}
  \label{fig:appredix-compare-occlusion2}
\end{figure*}

 \begin{figure}[tb]
  \centering
  \includegraphics[width=0.75\linewidth]{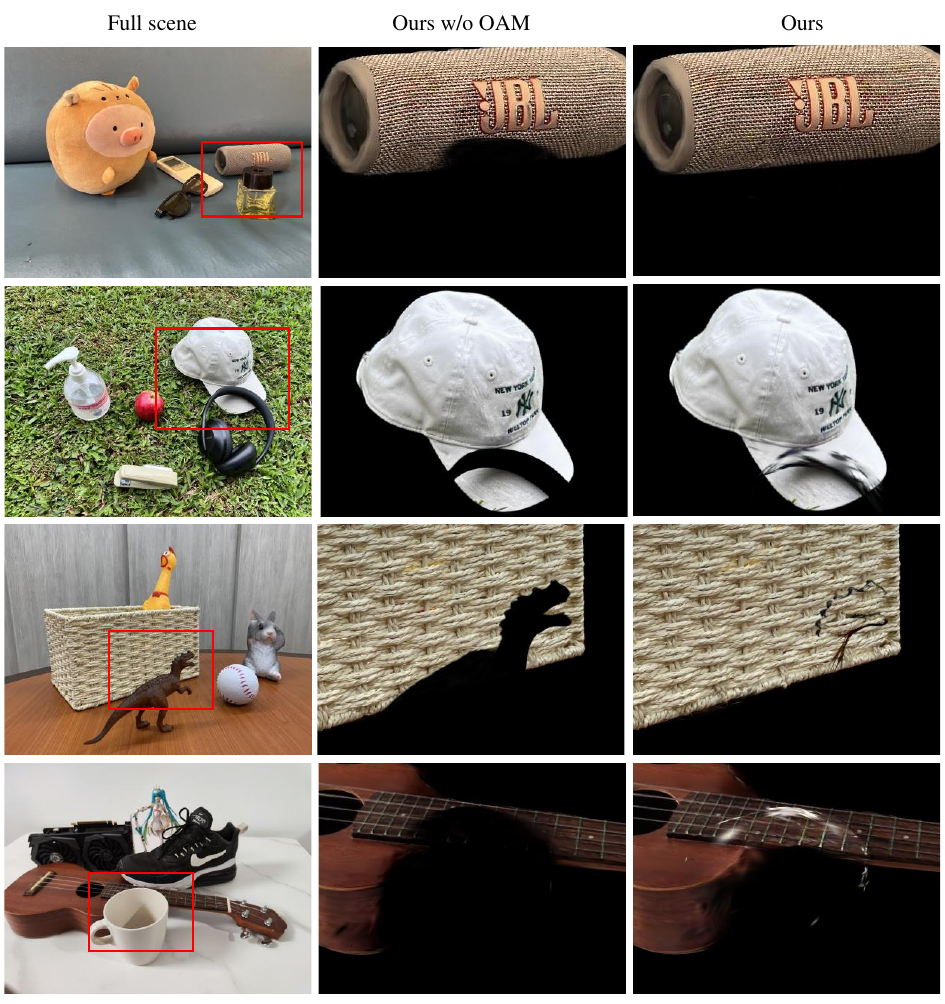}
  \caption{Ablation study of Occlusion Anlysis Model(OAM) on the 3D-OVS dataset.
  }
  \label{fig:compare-occlusion}
\end{figure}

 \begin{figure}[tb]
  \centering
  \includegraphics[width=0.8\linewidth]{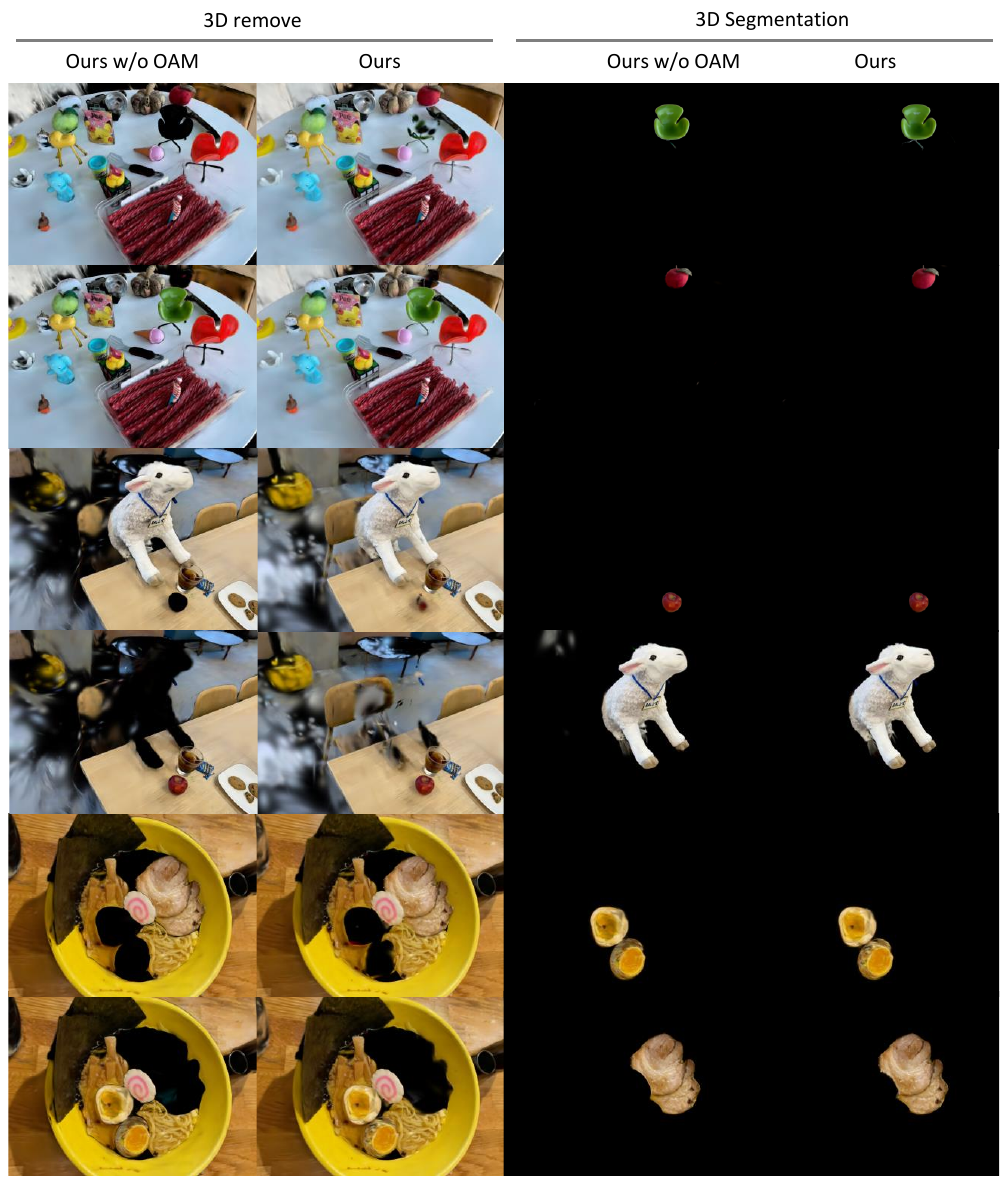}
  \caption{Ablation study of Occlusion Analysis Model(OAM) on the LERF-Mask dataset in remove and segmentation tasks.}
  \label{fig:2_seg_del}
\end{figure}

\subsubsection{Datasets and Metrics}
We show experiment results on four datasets. Two of them are face-forward scenes (3D-OVS~\cite{liu2023weakly} and LLFF~\cite{mildenhall2021nerf}), while the other two consist of 360$^\circ$ scenes(LERF-MASK~\cite{gaussian_grouping} and Mip-NeRF360~\cite{barron2022mip}). 
The 3D-OVS and LERF datasets are designed for open vocabulary 3D semantic segmentation, providing a comprehensive category list. 
We modified these datasets by removing occluded segmentation masks, i.e. background masks. 
For avoiding ambiguity, we select one mask of the object as the 3D segmentation prompt, while masks from other views are used as the test set. 
We render the extracted 3D Gaussians in test views and compare with ground truth image. 
The Peak Signal-to-Noise Ratio~(PSNR) is used as the evaluation metric to reflect the fidelity of the 3D segmentation results.
In addition, we extended the LLFF and Mip-NeRF360 datasets by adding masks of salient objects for evaluation.
The number of training and test images for each dataset is listed in Table~\ref{tab:dataset}.

LangSplat and Feature-3DGS support 3D segmentation by filtering out Gaussians with relevancy scores lower than a chosen threshold. To choose the best threshold, we crop the rendering images by the bounding box of the prompt mask, try 1000 score thresholds from 0 to 10 with a 0.01 increment at each step, and select the threshold with the best PSNR.
We identify the best threshold for each prompt, respectively.
Gaussian grouping supports 3D segmentation by a selected object ID.
We identify the object ID by finding the mask that overlays mostly with the prompt mask.

\subsubsection{Implementation Details}
To obtain consistent mask labels, we use DEVA to track the objects from training images.
We start including label loss from the 1000th iteration.
To reduce the GPU memory cost and improve efficiency, we randomly sample 10 object masks in an image for optimization at each iteration when computing label loss.
In the process of lifting labels to 3D Gaussians, we assign the label of a pixel to a Gaussian only if the Gaussian's contribution to the pixel exceeds a threshold of 0.6.
For scenes with a resolution of 1440$\times$1080, the model training on a single 3090 GPU takes approximately 23 minutes for 15,000 iterations.

\subsection{Quantitative Results}

\subsubsection{Description}
We compared our method with other SOTA methods on the four datasets: 3D-OVS, LLFF, LERF-MASK and Mip-NeRF360, as detailed in~\Cref{tab:comparison}. The experiments on Feature-3DGS only conducted on one dataset, due to its long training time.
The table shows that our method extracts 3D Gaussian more clearly the four datasets.
Our result significantly surpasses the Feature-3DGS~\cite{zhou2024feature} with 512-dimensional features, LangSplat~\cite{qin2023langsplat} with autoencoder and also exceeds the Gaussian Grouping~\cite{gaussian_grouping} with 16-dimensional identity features, demonstrating the superiority of our method.
SSIM and LPIPS results for four datasets are shown in Tabel~\ref{tab:comparison_ssim_lpips}. 
Our method slightly outperforms SAGS and significantly surpasses the Feature-3DGS, LangSplat and Gaussian Grouping.
We also reported the training time on an NVIDIA RTX-3090 GPU in~\Cref{tab:speed}. Our method achieves a 22$\times$ speed improvement over Feature-3DGS, highlighting the significant efficiency of our method.

\subsubsection{Analysis}
Taking the result on 3D-OVS dataset as an example, Feature-3DGS achieves a higher score than LangSplat. This is because Feature-3DGS learns a full 512-dimension feature for each 3D primitive Gaussian,  while LangSplat only learns a compressed 3-dimensional feature for each Gaussian. This compression may severely impact the semantic feature learning capability of LangSplat. 
Gaussian Grouping achieves a higher mIoU of extracted Gaussian compared to Feature-3DGS, indicating that grouping operation during optimization helps decouple Gaussians representing different objects. 
Finally, Our method achieves the best performance, revealing that directly assigning the group-ID or label to the main rendering Gaussians is more effective and computationally efficient than assigning the group-ID using color alpha-bender weights approaches, including Gaussian Grouping.


\subsection{Qualitative Results.}
To illustrate the segmentation performance of the 3D Gaussians, we rendered the segmented 3D Gaussians from the test view in Figure~\ref{fig:compare}.
It is evident that other methods fail to effectively disentangle the Gaussians to represent different objects, often resulting in the inclusion of Gaussians from unrelated objects.
In contrast, our method accurately extracts Gaussians with precise boundary representations. 
Additional qualitative results are provided in the appendix.


\subsection{Ablation Studies}
We conducted ablation experiments on four datasets(3D-OVS, LLFF, LERF, Mip-NeRF360), as show in Table~\ref{tab:comparison}. 
3D-OVS images are sparse views while other three dataset images are dense enough to obtain consistent cross-view mask without Densification of View(DOV). So we only conduct w/o and w/ DOV ablation study on 3D-OVS dataset. DOV improves segmentation accuracy on sparse view datasets like  3D-OVS. 

Gaussian Projection Filter(GPF) improves the face-forward dataset 3D-OVS and LLFF, while have negative affect on 360 degree dataset LERF and Mip-NeRF360. We guess 360 degree dataset provide enough visual constraint to resolve Gaussian label conflict and GPF as a additional regularity may slightly damage the reconstruction performance.

Occlusion Analysis Model(OAM) improves the 3D reconstruction of occluded region, as shown in Fig. 7. The occluded part must be visible from other view images so that our method may learn its appearance. OAM will keep the learned object appearance in the reason spatial position.
In 3D remove and 3D segmentation task, OAM presents more reasonable background in 3D remove task and no obvious degeneration on 3D segmentation task, shown in Fig~\ref{fig:2_seg_del}.
Since the test set are almost all occlusion-free, so the performance  of OAM in Table~\ref{tab:comparison} are changed slightly.

\section{Limitation}

LabelGS learns consistent cross-view labels generated by DEVA. This means that LabelGS may struggle to segment objects that do not covered by these masks. Improving LabelGS's generalization ability to DEVA-tracking-failed targets is a promising direction for future research.
Due to the introduction of constraints during optimization, the peak signal-to-noise ratio (PSNR) of all Gaussian model renderings exhibits a slight decline, a phenomenon analogous to that observed in the Gaussian Grouping method \cite{gaussian_grouping}.

\section{Conclusion}

This paper presents LabelGS, a method for constructing labeled 3D Gaussian models that support precise and efficient 3D Gaussian segmentation.
We use a video tracking module to obtain consistent cross-view labels.
The label of a pixel is then assigned to the main Gaussian in rendering, building the labeled Gaussian model.
To resolve occlusion overfitting, we employ Occlusion Analysis Model(OAM) to exclude the occluded areas from loss computation, achieving accurate 3D representations.
In face-forward datasets, we propose Gaussian Projection Filter(GPF) to improve the accuracy of 3D segmentation.
Experiments have shown that our method outperforms SOTA methods such as Feature-3DGS, with a 12.7\% increase in PSNR and a 22$\times$ acceleration in speed.

\subsubsection{Acknowledgement.}

This work was supported by National Key R\&D Program of China (Grant No. 2024YFF0618400), National Natural Science Foundation of China (Grant No. 62320106007) and Guangdong Provincial Key Laboratory (Grant No.\allowbreak 2023B1212060076), and ZTE Industry-University-Institute Cooperation Funds under Grant (No.\allowbreak IA20241202008).

%
%
%
\bibliographystyle{splncs04}
\bibliography{main}

\end{document}